# Predicting Cancer Using Supervised Machine Learning: Mesothelioma

**Avishek Choudhury**

*School of Systems and Enterprises, Stevens Institute of Technology, Castle Pointe, Hoboken, 07030, New Jersey, USA.

achoudh7@stevens.edu
Orcid: https://orcid.org/0000-0002-5342-0709

**Abstract.**

**Background:** Pleural Mesothelioma (PM) is an unusual, belligerent tumor that rapidly develops into cancer in the pleura of the lungs. Pleural Mesothelioma is a common type of Mesothelioma that accounts for about 75% of all Mesothelioma diagnosed yearly in the U.S. Diagnosis of Mesothelioma takes several months and is expensive. Given the risk and constraints associated with PM diagnosis, early identification of this ailment is essential for patient health.

**Objective:** In this study, we use artificial intelligence algorithms recommending the best fit model for early diagnosis and prognosis of MPM.

**Methods:** We retrospectively retrieved patient's clinical data collected by Dicle University, Turkey and applied multilayered perceptron (MLP), voted perceptron (VP), Clojure classifier (CC), kernel logistic regression (KLR), stochastic gradient decent SGD), adaptive boosting (AdaBoost), Hoeffding tree (VFDT), and primal estimated sub-gradient solver for support vector machine (s-Pegasos). We evaluated the models, compared and tested using paired T-test (corrected) at 0.05 significance based on their respective classification accuracy, f-measure, precision, recall, root mean squared error, receivers' characteristic curve (ROC), and precision-recall curve (PRC).

**Results:** In phase-1, SGD, AdaBoost. M1, KLR, MLP, VFDT generate optimal results with the highest possible performance measures. In phase 2, AdaBoost, with a classification accuracy of 71.29%, outperformed all other algorithms. C-reactive protein, platelet count, duration of symptoms, gender, and pleural protein were found to be the most relevant predictors that can prognosticate Mesothelioma.

**Conclusion:** This study confirms that data obtained from Biopsy and imagining tests are strong predictors of Mesothelioma but are associated with a high cost; however, they can identify Mesothelioma with optimal accuracy.

**Keywords:** Mesothelioma, Predictive modeling, Decision support system, Machine learning, Artificial intelligence, Lung cancer,

# 1. Background

Pleural Mesothelioma (PM) is an unreceptive tumor of mesothelial cells associated with prior exposure to asbestos. The 2015 World Health Organization classification subdivides mesothelioma tumors into three histological categories: (a) epithelioid, (b) biphasic, and (c) sarcomatoid MM [1]. Despite the availability of chemotherapy [2, 3] and a diverse range of clinical inspections, accurate prognostication of PM has been a concern among clinicians and patients. PM is an exceptionally unique ailment. Its staging system [4] results in a confusing preliminary identification process [5, 6], and the different biology [7] deters accurate prognostication. Within a general population, PM affects about 2 people per million per annum [8]. Additionally, due to more exposure to asbestos [9], industrialized zones are severely affected by PM [9-11]. It has been estimated that death due to Mesothelioma in Western Europe will almost double every year. About 9000 deaths were estimated around 2018, and a projection of a quarter of a million deaths is calculated by 2029.

The severity of Mesothelioma can be categorized into stage 1, stage 2, stage 3, and stage 4 (cancer). Stage1 and stage 2 symptoms of PM, such as dry coughing, dyspnea, respiratory complications, chest or abdominal pain, fever, pleural effusion, fatigue, and muscle weakness are very weak predictors of Mesothelioma [12]. Since Mesothelioma is rare, patients are less likely to be suspected of the disease. Moreover, its initial symptoms during stage 1 and 2 are similar to common diseases such as pneumonia and irritable bowel syndrome [13], PM can also be misdiagnosed as adenocarcinoma which is a non-terminal lung cancer [13]. If Mesothelioma is not diagnosed and meets no medical aid at its early stage, it rapidly burgeons into a phase 3 or stage 4 cancer. Unfortunately, the survival rate after being diagnosed with last-stage Mesothelioma is typically about a year. To treat Mesothelioma effectively, an early diagnosis is recommended.

Diagnosing Mesothelioma is challenging, and the expenses associated with identifying this disease can ascend rapidly. In fact, since the primary way to diagnose Mesothelioma incorporates ruling out other plausible diseases, more frequently than not, many examinations may be administered that isn't exclusive to Mesothelioma itself but are for erstwhile disorders instead [14]. Furthermore, it is often suggested getting a second opinion [14], recapping many of the diagnostic tests over and over. For all these causes, diagnostic expenses for Mesothelioma starts piling up even before the necessary treatment commences. Mesothelioma diagnosis typically implicates taking imaging scans of tumors, examining a biopsy of cancer tissue, and blood tests [13].

Existing Mesothelioma diagnosis employs multiple imaging tests, such as X-rays, CT scans, MRI, and PET scan [13], all of which are expensive. The specialized imaging equipment is costly both for an upfront purchase and for maintenance. Secondly, this equipment requires well-trained technicians to ensure apt operation of the device. A patient can presume to spend about $800 — $1,600 [14] for a single CT, MRI, or PET scan, respectively. Moreover, multiple scans may be required during diagnosis [14], which can quickly burgeon the overall costs.

Among all the existing means of diagnosis, Biopsy has been recognized as the most accurate invasive method that confirms Mesothelioma [13]. It is a procedure that requires the removal of fluid or tissue samples from the tumor or cancer site and their analysis under a microscope. There are many diverse approaches to obtaining a biopsy, and which one to be used depends on the suspected tumors' location. Some biopsies embrace making an incision and inserting implements to obtain a sample of the tumor cell, while others only require a needle. Given the wide range of biopsy procedures, its expenses can range from $500 to $700 for a needle biopsy [14], $3,600 to $5000 for pleuroscopy (lungs) or laparoscopy (abdomen) [14], $7,800 to $7,900 for thoracotomy

(lungs) or laparotomy (abdomen) [14]. Like other diagnostic procedures, biopsies may also require to be performed multiple times [14], increasing the overall diagnosis expenses. Doctors also explore a variety of blood tests such as MESOMARK, SOMAmer, and Human MPF to look for biomarkers that suggest Mesothelioma [13]. However, currently, no blood tests are precise enough to confirm a diagnosis on their own [13].

Malignant Pleural Mesothelioma has the potential to grow into cancer and sabotage patient health. Like any other fatal disease, PM demands early diagnosis and effective treatment. However, effective diagnosis methods, such as thoracotomy and pleuroscopy, are costly and might not be affordable for patients worldwide [15, 16]. Additionally, about two-thirds of the world do not have adequate access to the required technologies, expensive imaging devices, and expert technicians [17].

There exists some work of literature that has used artificial intelligence algorithms such as but are not limited to a decision tree, random forest, support vector machine, and artificial neural network to identify PM [18] but with some limitations. Decision tree models, including Random Forest, are prone to overfitting [19] or fails to generate 100% accuracy or might also fail to converge a large data set [20].

In our study, we propose a model that overcome the flaws as mentioned above and can diagnose PM with and without requiring data from expensive Biopsy and imaging tests.

## 2. Methods

In this study, we used data extracted from clinical reports generated by Dicle University, Turkey. The data set consists of 324 patient data with 35 variables. Out of 324 observations, 41% were females. The patients involved in this study were in nine different cities. We performed k-fold cross-validation to minimize any bias and variance in the data set. Cross-validation is a resampling technique used to gauge machine learning models on a limited data set. In this method, we divided the data set into a K-folds and used each fold as a testing set (Figure 1).

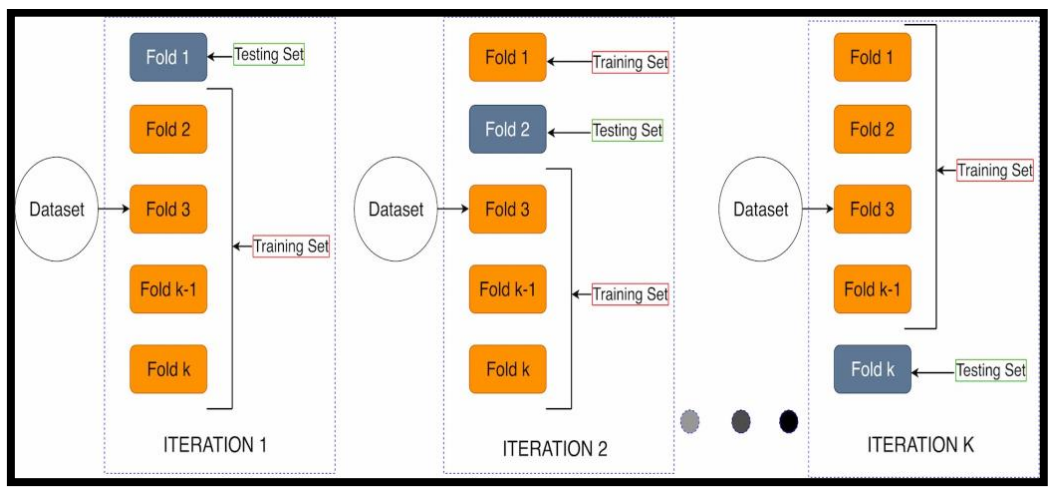

**Fig. 1**. *K-fold Cross-Validation*

In this study, we considered the value of k to be 10 becoming 10-fold cross-validation. The selection of k is usually 5 or 10 [21], and this value of k prohibits extreme high bias and variance

[22]. All the analysis was performed using R-studio, an open-source machine learning and statistical tool, and Waikato Environment for Knowledge Analysis (WEKA), a free software suite of machine learning licensed under the GNU General Public License, programmed in JAVA, and developed at the University of Waikato, New Zealand.

Table 1 lists all the attributes contained in our data set, and it also determines the mean, deviation, and logistic correlation of all predictors with the target variable ("class of diagnosis"). The data set used in this study had an imbalanced class variable, as shown in Figure 2. Consecutively, Figure 3 exhibits the correlation matrix of the dataset, and Figure 4 shows the numeric variables, respectively.

**Table 1.** *Data Statistics*

| Predictor | Mean | Deviation | Logistic correlation ("class of diagnosis") |
|---|---|---|---|
| Age | 54.74 | 11.00 | 0.06 |
| Gender | - | - | 0.15 |
| City | NA | NA | 0.02 |
| Asbestos exposure | 0.86 | 0.34 | 0.07 |
| Type of MM | 0.05 | 0.26 | 0.13 |
| Duration of asbestos exposure | 30.18 | 16.41 | 0.06 |
| Diagnosis method* | - | - | 1.00* |
| Keep side | 0.75 | 0.56 | 0.10 |
| Cytology | 0.28 | 0.45 | 0.02 |
| Duration of symptoms | 5.44 | 4.71 | 0.02 |
| Dyspnea | 0.81 | 0.38 | 0.02 |
| Ache on chest | 0.68 | 0.46 | 0.05 |
| Weakness | 0.61 | 0.48 | 0.06 |
| Habit of cigarette | 0.91 | 1.15 | 0.05 |
| Performance status | 0.52 | 0.50 | 0.03 |
| White blood | 9457.45 | 3450.73 | 0.05 |
| Cell count (WBC) | 9.55 | 3.34 | 0.05 |
| Hemoglobin (HGB) | 0..42 | 0.49 | 0.03 |
| Platelet count (PLT) | 369.65 | 227.55 | 0.06 |
| Sedimentation | 70.68 | 21.74 | 0.00 |
| Blood lactic dehydrogenize (LDH) | 308.91 | 185.14 | 0.01 |
| Alkaline phosphate (ALP) | 66.16 | 35.07 | 0.04 |
| Total protein | 6.58 | 0.82 | 0.01 |
| Albumin | 3.30 | 0.63 | 0.04 |
| Glucose | 112.41 | 38.46 | 0.01 |
| Pleural lactic dehydrogenize | 518.47 | 536.27 | 0.03 |
| Pleural protein | 3.93 | 1.57 | 0.03 |
| **Pleural albumin** | 2.07 | 0.91 | 0.07 |
| **Pleural glucose** | 48.44 | 27.23 | 0.01 |
| **Dead or not** | - | - | - |
| **Pleural effusion** | 0.87 | 0.33 | 0.03 |
| **The pleural thickness on tomography** | 0.59 | 0.49 | 0.01 |
| **Pleural level of acidity (pH)** | 0.52 | 0.50 | 0.04 |
| **C reactive protein (CRP)** | 64.18 | 22.66 | 0.11 |

Mesothelioma data set can be broadly divided into pre-diagnosis data and post-diagnosis data. Pre-diagnosis data refers to the all the records obtained before Mesothelioma was clinically confirmed such as patient age, gender, the city they belonged to, smoking habit, exposure to asbestos, duration of exposure to asbestos, early-stage symptoms including the feeling of weakness, heartache, and dyspnea, and duration of symptoms. Pre-diagnosis data also encompasses blood test results such as white blood cell count, hemoglobin level, platelets count, and others. Post-diagnosis are those data that refers to the records retrieved after Mesothelioma was confirmed. Type of Mesothelioma detected (a type of MM), side effects of chemotherapy (keep side), and survival of the patient after treatment (dead or not) are all post-diagnosis data. This study eliminates the "dead or not" predictor from all analyses.

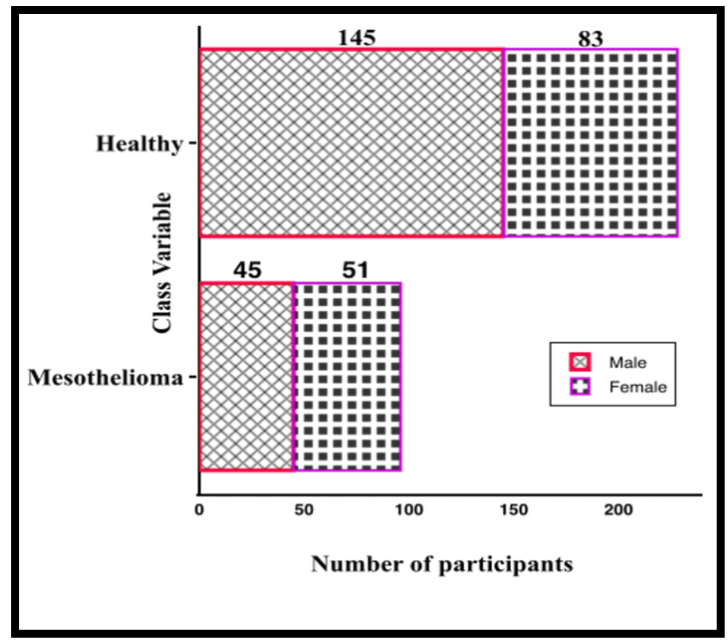

**Fig. 2.** *Unbalanced class variable. The data used was significantly unbalanced with an Odd ratio of 0.5062 and a p-value of 0.0065 at a 95% confidence interval.*

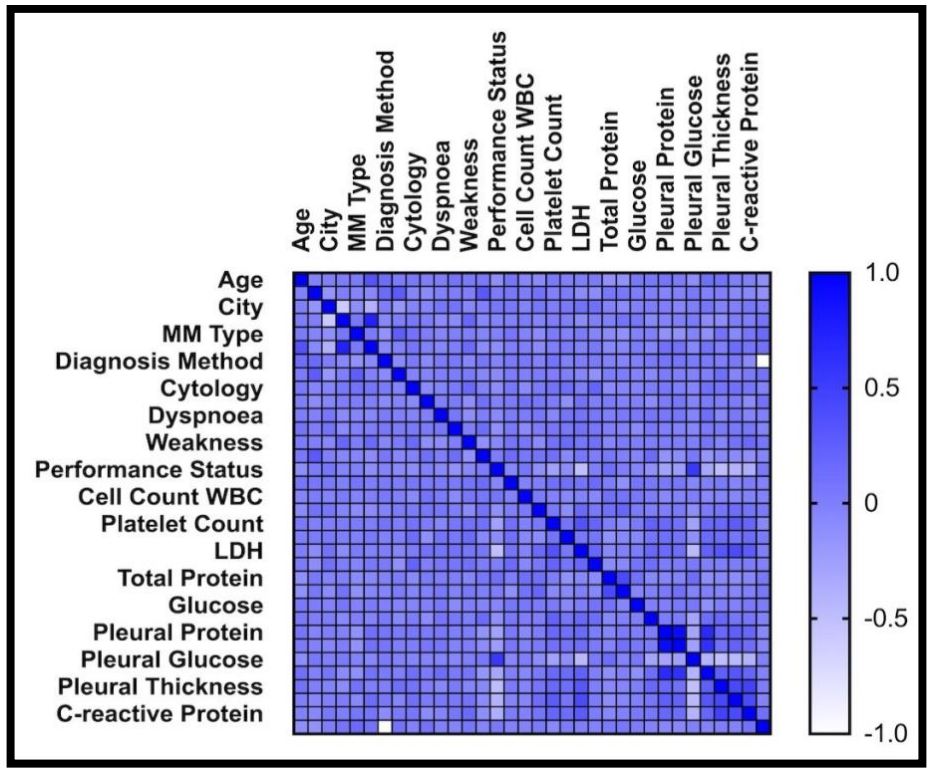

**Fig. 3.** *Correlation Matrix*

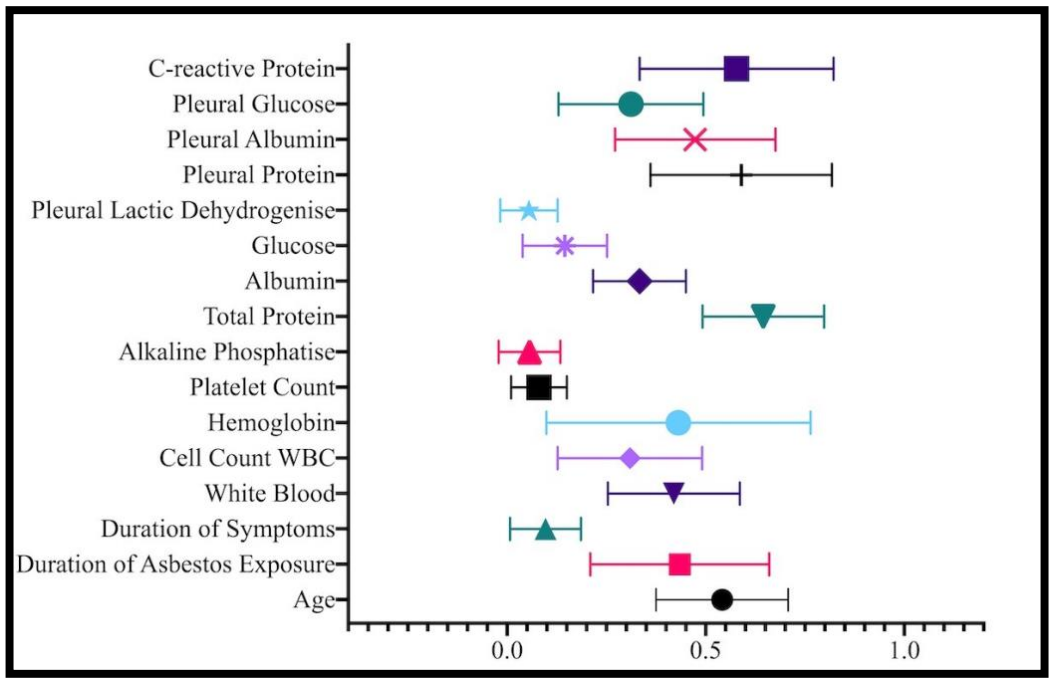

**Fig. 4.** *Normalized representation of the numeric variables.*

The predictor "diagnosis method" refers to data obtained from invasive Biopsy and imaging test results. Invasive Biopsy and imaging tests can accurately identify Mesothelioma but are expensive procedures and may require repeated examinations, as stated earlier. To advocate the applicability of AI predictive analytics on both pre and post-diagnostic data, we perform a comparative analysis of classification models into two phases. Phase-1 models use all the predictor variables except "dead or not" as input to produce high classification accuracy. The same set of models in Phase-2 only takes relevant predictors from pre-diagnosis data as its input.

Phase-1 and phase-2 are denoted as *high accuracy* and *low-cost* phases respectively because phase-1 execution demands data from expensive, invasive Biopsy and imaging test results, which are robust predictors of MM (logistic correlation = 1) and thus the model is expected to yield high accuracy. Whereas, phase-2 considers only predictors with lower logistic correlation (pre-diagnosis data) and eliminates the use of invasive Biopsy and imaging test results. Execution of phase-2 also incorporates a feature selection method to enhance its accuracy and reduce computational time.

Healthcare data sets are typically designated with several predictors for effective model structure [23]. Commonly majority of these predictors are extraneous to the classification, and perceptibly their significance is unknown in advance [23]. Many difficulties arise while dealing with large feature sets. One is affably technical —large data sets impedes computational speed, consumes resources, and is merely bothersome. Another is even more vital — many artificial intelligent algorithms reveal a diminution of performance when the number of variables is considerably higher than optimal. Therefore, the selection of the smallest feature set that can yield the best possible classification outcome is recommended. This problem also coined as the minimal-optimal problem [24] and has been studied extensively. Moreover, many algorithms have been established to reduce the feature set to a manageable and optimal size.

Nevertheless, the problem mentioned above follows another setback —the "*all-relevant problem*" [23]. Determining only the significant attributes is beneficial for developing predictive models. However, it is essential to consider all attributes to understand the fundamental mechanisms of the subject of interest. For instance, in an outpatient clinic, listening to the patient's concern, preferences, family support is as important as listening to their medical history and symptoms before making any clinical decision. However, to build a predictive model, only data related to the disease is required. An extensive discussion defining the importance of finding relevant attributes is reported by Nilsson et al. in 2007 [24].

In phase-2 of our study, we used the Boruta algorithm and identified all relevant predictors. Boruta algorithm is a wrapper built around the random forest algorithm [25]. Boruta uses Z-score to measure the importance of a variable. Since we cannot use Z-score staunchly to determine variable relevance, an external reference is used to confirm the essence of all attributes. Boruta creates a 'shadow variable' by shuffling values of the original attribute then performs a classification using the shadow and original variables to determine the importance of all predictors.

The following algorithms were implemented, compared, and tested using *paired T-test* (*corrected*) at 0.05 significance.

## 2.1. Algorithms

### 2.1.1. Stochastic Gradient Descent (SGD)

Gradient descent is a method to determine the local minima. Stochastic gradient descent is gradient descent performed using multiple updates at a time on a small batch (mini batch) of the data set selected at random (stochastically). Instead of calculating the gradient of the cost (error) based on the whole data set, SGD breaks the data set into mini-batches and compute the gradient on each batch separately, followed by a neural net update based on the partial gradient. In other words, it is an optimization algorithm that iteratively determines the values of learnable parameters of a function (f) to minimize the cost function (error rate). The cost function for our study is a root mean squared error, which can be determined using the following equation (Eq. (1)).

$$RMSE = \sqrt{\frac{1}{N}\sum_{i=1}^{n}(y_i - (mx_i + b))^2} \tag{1}$$

Mathematically, SGD is a simplification of gradient descent. Instead of calculating the gradient of $E_n(f_w)$ (empirical risk using gradient descent), each iteration estimates this gradient by a single randomly picked example (Eq. (2)):

$$z_t: w_t - \gamma_t \nabla_w Q(z_t - w_t) \tag{2}$$

Where z is a random pair of input x and scalar output y; w is weight; $\gamma$ is learning rate; $Q(z, w)$ is the loss.

### 2.1.2. Adaptive Boosting M1

It is also known as AdaBoost. M1 is a meta-estimator applied concurrently with other weak algorithms to enhance performance. AdaBoost can fine-tune the weak algorithms in favor of misclassified instances by previous classifiers. AdaBoost-M1 can be mathematically defined as given below (Eq. (3)).

The weak learner accepts input $x$ to yield a value indicating the class of the object. Each weak learner produces an output hypothesis, $h(x_i)$. At each iteration $t$, a weak learner is generated and assigned a coefficient $t$ such that the sum of training error $E_t$ (Eq. (4)) of the resulting t-stage boost classifier is minimized.

$$E_t = \sum E|F_{t-1}(x_i) + \alpha_t h(x_i) \tag{4}$$

Where $F_{t-1}(x)$ is the boosted classifier from the previous stage of training. $E(F)$ is an error function, and $f_t(x) = \alpha_t h(x)$ is the weak learner.

### 2.1.3. Kernel Logistic Regression (KLR)

This algorithm uses the kernel to allow classification of linearly non-separable. The kernel is a conversion function that satisfies mercer's necessary settings, stating that a kernel function must be denoted as an inner product and be positive semi-definite.

### 2.1.4. Multi-layer Perceptron

It is a type of Artificial Neural Network (ANN). It typically consists of input, hidden, and an output layer. The input layer contains set of neurons $\{x_i | x_1, x_2, ..., x_m\}$ signifying the input variables. Every neuron in the hidden layer converts the input with a weighted linear summation ($w_1x_1 + w_wx_2 + \cdots + w_mx_m$), Followed by an activation function. The output layer collects the values from the hidden layer and yields the result.

Figure 5 shows a standard neuron model with two parts. In the first part, the input data are assembled for a sum. Each weight ($w_i$) compeers a data dimension ($x_i$), while bias ($w_0$) adds to the intercept of the function [26].

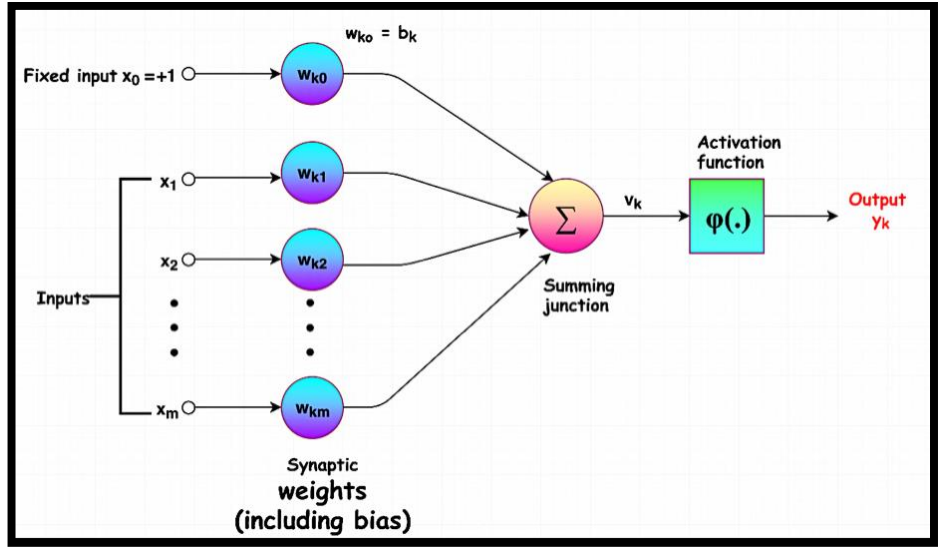

**Fig. 5.** *Typical Neural Network.*

In the second half, the initiation function uses the activation function such as ReLU, Sigmoid, and Tanh [26] to obtain a nonlinear eigenvalue. MLP learns a function $f(.)\colon R^m \to R^o$ by training a data set, where *m* is the input dimension, and *O* is the dimension of the output. Figure 6 shows a 4 layered (L1 through L4) neural network, where L1 represents the input layer, L4 the output layer, and L2, L3 the hidden layers.

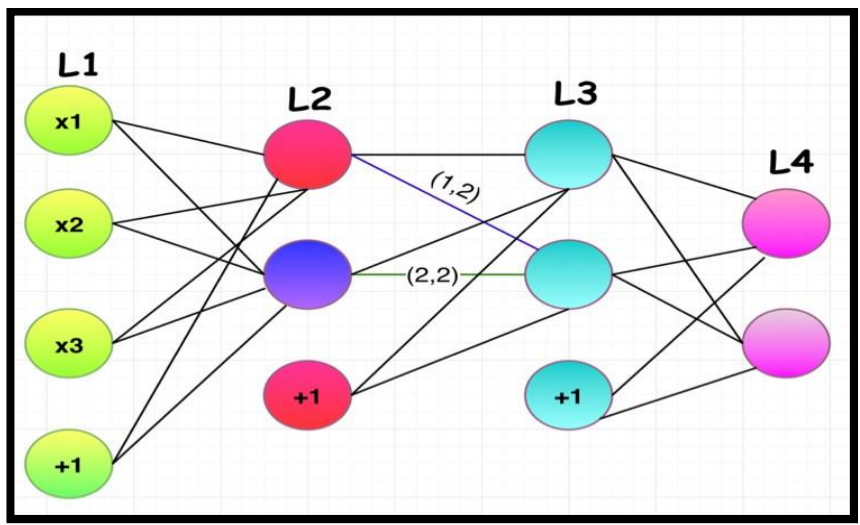

**Fig. 6.** *Feed-forward neural network.*

### 2.1.5. Voted Perceptron (VP)

This algorithm joins the Rosenblatt's perception and Helmbold-Warmuth's leave-out method to calculate the weight vectors which during the learning process vote on a prediction. The output of voted perceptron is given by Eq. (6):

$$y_i = sign(\sum_{p=0}^{P} c_p \, sign(w_p, x)) \qquad (6)$$

Where $x$ inputs, $p = 0,1,2,...P$; $w_p$ are weights, $y_i$ is the predicted class, and $c_p$ is the survival time (reliability of $w_p$).

### 2.1.6. Hoeffding Tree

It is also known as Very Fast Decision Tree (VFDT) is a tree algorithm for data stream classification. The Hoeffding tree is an incremental decision tree learner for a large data set, that assumes that the data distribution is constant over time. It grows a decision tree based on the theoretical guarantees of the Hoeffding bound. In other words, VFDT employs Hoeffding bound to calculate the minimum number of observations to reach a desirable level of confidence in splitting the node.

### 2.1.7. Clojure Classifier (CC)

It is a wrapper classifier developed in Clojure programming language. It mandates to have at least a learn-classifier function and distribution-for-instance function. The learn-classifier function takes an object and a string (nullable) and returns the learned model as a serializable data structure. The distribution for-instance function takes an instance to be predicted and a model as an argument and returns the prediction as an array.

### 2.1.8. Primal Estimated sub-Gradient Solver for SVM

It is also known as s-Pegasos. It performs SGD on a primal objective (Eq. (7), Eq. (8)) with carefully chosen to step size.

$$\min_{w} \frac{\lambda}{2}|W|^2 + \frac{1}{m} \sum_{(x,y) \in S} l(W;(X,y)) \qquad (7)$$

Where

$$l(W;(X,y)) = \max(0, 1 - y(w,X)) \qquad (8)$$

### 2.2. Model evaluation

Supervised machine learning models are primarily evaluated based on their classification accuracy. However, there are other crucial performance measures such as f-measure, root mean squared error (RMSE), receiver operating characteristic (ROC), and precision-recall curve (PRC).

Classification accuracy is the metric for evaluating classification models. It is the fraction of predictions or classification that a model performs correctly. Classification accuracy can be calculated by the given equation (Eq. (9))

$$Accuracy = \frac{Number\ of\ correct\ prediction}{Total\ number\ of\ prediction} \tag{9}$$

$$= \frac{TP + TN}{TP + TN + FP + FN}$$

Where TP = True positive; TN = True negative; FP = False positive; FN = False-negative. The AUC-ROC curve is used as a classification performance measurement at different threshold settings. It is a graphical representation sensitivity and false-positive, as shown in Eq. (10).

$$FPR = (1 - specificity) \tag{10}$$

Regarding information retrieval undertakings with binary classification (relevant or not relevant), precision is the segment of retrieved instances that are relevant. In contrast, recall, also known as sensitivity, is the fraction of retrieved cases to all relevant situations. In this context of information retrieval, the PRC becomes very useful. PRC is a graphical representation of recall (x-axis) and precision (y-axis), where recall and precision are determined using the given formula (Eq. (11), Eq. (12)), respectively.

$$Recall = \frac{TP}{TP + FN} \tag{11}$$

$$Precision = \frac{TP}{TP + FP} \tag{12}$$

F-measure, also known as F1-score, is the harmonic mean of precision and recall (Eq. (13)), where f-measure reaches its best at 1 and worst at 0.

$$f1\ score = \frac{2 * (Precision * Recall)}{Precision + Recall} \tag{13}$$

The root-mean-square error (RMSE) is a measure of the performance of a model. It does this by computing the difference between predicted and the actual values as given below Eq. (14).

$$RMSE = \sqrt{\sum_{i=1}^{N} \frac{(x_i - y_i)^2}{N}} \tag{14}$$

Where $(x_i - y_i)$ is the difference between the predicted and actual value, and N is the sample size.

## 3. Results

### 3.1. Phase 1

As shown in Table 2, SGD, AdaBoost. M1, KLR, MLP, VFDT generates perfect results with 100% accuracy, precision, recall, and f-measure. These algorithms also return the highest possible ROC, PRC, and zero RMSE. s-Pegasos also delivers close to the optimal result. In this phase, the high accuracy of 100% indicates that results obtained from Biopsy and imaging tests are very strong predictors of MM. This result validates the significance of biopsy and imaging results ("diagnosis method") from a data science viewpoint.

Table 2. *Comparing AI performance (phase 1)*

| Algorithm | SGD | AdaBoost.M1 | KLR | MLP | VP | VFDT | CC | s-Pegasos |
|---|---|---|---|---|---|---|---|---|
| Classification accuracy (%) | 100 | 100 | 100 | 100 | 70.38 | 100 | 70.38 | 99.36 |
| f-measure | 1.00 | 1.00 | 1.00 | 1.00 | 0.83 | 1.00 | 0.83 | 1.00 |
| Recall | 1.00 | 1.00 | 1.00 | 1.00 | 1.00 | 1.00 | 1.00 | 1.00 |
| Precision | 1.00 | 1.00 | 1.00 | 1.00 | 0.70 | 1.00 | 0.70 | 0.99 |
| ROC | 1.00 | 1.00 | 1.00 | 1.00 | 0.50 | 1.00 | 0.50 | 0.99 |
| PRC | 1.00 | 1.00 | 1.00 | 1.00 | 0.70 | 1.00 | 0.70 | 0.99 |
| RMSE | 0.00 | 0.00 | 0.00 | 0.00 | 0.54 | 0.01 | 0.54 | 0.04 |

Phase 2 demonstrates the relevance of pre-diagnosis data. It also shows the behavior of all predicting models post removal of "diagnosis method" and other post-diagnosis data.

### 3.2. Phase 2

Boruta algorithm confirmed five relevant attributes that are enough to predict the presence of Mesothelioma without any loss in the model's performance. In other words, the selected attributes alone can prognosticate MM with the same accuracy as all other pre-diagnosis predictors when taken together as input. The relevant predictor identified were c-reactive protein, platelet count, duration of symptoms, gender, and pleural protein.

This method neither downgrades the remaining predictors nor does it recommend revising the regular clinical procedures. Figure 7 below shows the attributes recognized by the Boruta algorithm. Boruta plot generates a color-coded box plot for each attribute where green represents relevant predictors and red otherwise. The x-axis represents each of the candidate explanatory variables. The green box plots refer to the relevant attributes, whereas the red ones are identified as unimportant (from a data science viewpoint). Table 3 shows the performance measures of all AI algorithms employed in this study.

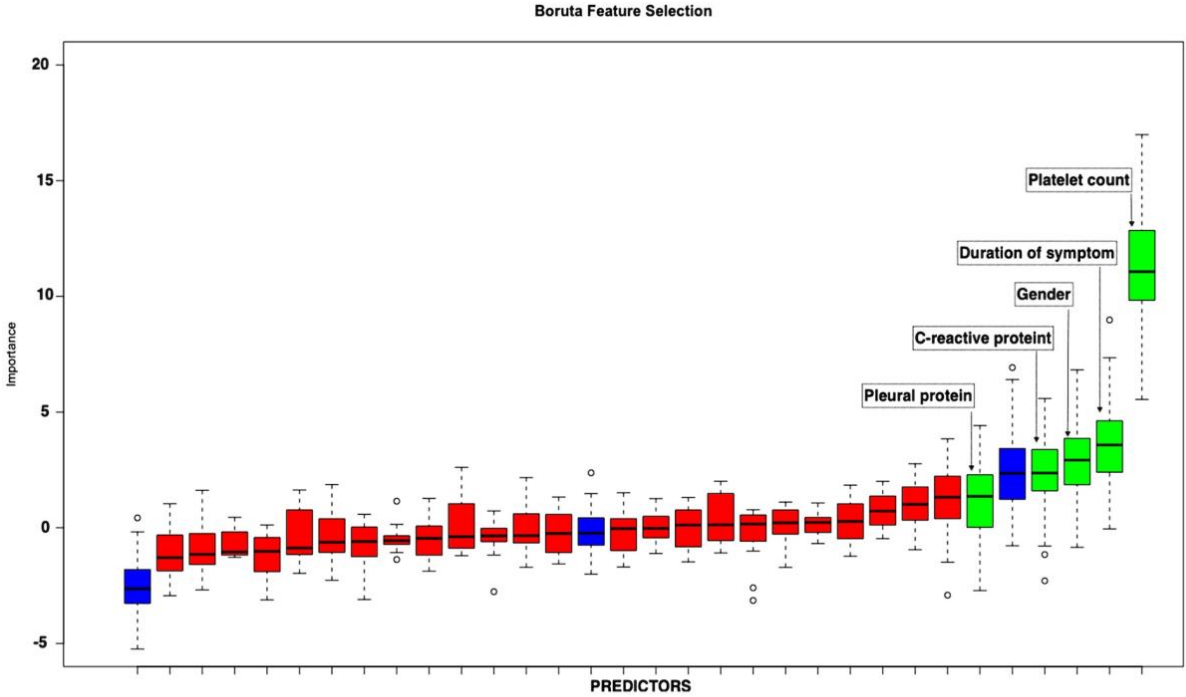

**Fig. 7.** *Boruta plot for feature selection*

AdaBoost outperformed all other models with the highest classification accuracy of 71.29%. Excluding the "diagnosis method" from the prediction model resulted in decreased accuracy. However, this phrase has an advantage. Despite lower accuracy, phase-2 helps to reduce diagnostic expenses.

**Table 3.** *Comparing AI performance (phase 2).*

| Algorithm | SGD | AdaBoost.M1 | KLR | MLP | VP | VFDT | CC | s-Pegasos |
|---|---|---|---|---|---|---|---|---|
| Classification accuracy (%) | 69.23 | 71.29 | 69.51 | 64.11 | 70.38 | 70.38 | 70.38 | 67.03 |
| f-measure | 0.80 | 0.82 | 0.79 | 0.74 | 0.83 | 0.83 | 0.83 | 0.77 |
| Recall | 0.80 | 0.86 | 0.93 | 0.84 | 1.00 | 1.00 | 1.00 | 0.81 |
| Precision | 0.74 | 0.74 | 0.76 | 0.75 | 0.70 | 0.70 | 0.70 | 0.75 |
| ROC | 0.58 | 0.61 | 0.65 | 0.61 | 0.50 | 0.50 | 0.50 | 0.58 |
| PRC | 0.74 | 0.77 | 0.82 | 0.79 | 0.70 | 0.70 | 0.70 | 0.74 |
| RMSE | 0.55 | 0.45 | 0.46 | 0.57 | 0.54 | 0.46 | 0.54 | 0.57 |

## 4. Discussion

Our study presents a way to minimize incidentaloma. An incidentaloma (overdiagnosis) is an incidental finding that is diagnosed in an asymptomatic or symptomatic patient for an unrelated reason [27]. In other words, performing unnecessary testing until the actual diagnosis is performed. Identification of the Mesothelioma typically involves overdiagnosis, especially if the identified lesion is benign or an indolent tumor not likely to produce symptoms before the patient dies of another cause. Overdiagnosis leads to excessive testing that adds up healthcare expenses, anxiety in patients, and unnecessarily exposes the patient to potentially harmful radiations and invasive surgeries. Additionally, our study achieves some goals of the American College of Radiology (ACR) Incidental Findings Committee. The proposed method in this study fulfills the following:

- Reduce risks to patients from additional unnecessary examinations.
- Limit the costs of managing incidental findings to patients and the health care system.
- Achieve high prediction accuracy in recognizing Mesothelioma.
- The Boruta algorithm identified the most influential factors that can help clinicians make informed decisions-based evidence rather than incidental findings.

Diagnosis of Mesothelioma can be difficult, especially given that the disease is relatively uncommon. Moreover, Pleural fluid examination is not sensitive nor specific. CT-guided Biopsy, which established diagnosis in this case, is limited by small sample size. Give these challenges; an accurate diagnosis of PM is crucial at both the individual and public health levels. It has necessary medico-legal significance due to diagnosis-related compensation [28]. However, prognosticating PM is challenging due to its composite epithelial pattern and low likelihood of occurrence [28]. To advocate the prognosis of PM with high accuracy and low diagnostic cost, the current study designed and implemented a prediction model consisting of two phases. (phase 1 and 2).

To our knowledge, no previous studies have implemented our AI models and focused on reducing diagnosis expenses by eliminating biopsy and imaging test results from the data set. Phase-2 of our study proposes AdaBoost. M1 algorithm that can identify high-risk patients at lower cost by taking only blood test results and patient's demographic data. The outcome from phase-2 can provide the doctors with a list of high-risk patients. Doctors and other healthcare providers can then prescribe biopsy tests only to the identified patients for reconfirming PM using a phase-1 model with optimal accuracy. This approach will reduce unnecessary biopsy tests and thus reduce overall expenses by up to $7900 [14].

The recommended model (AdaBoost) in phase-2 requires c-reactive protein, platelet count, duration of symptoms, gender, and pleural protein as its input. The expenses to collect the required input data can range from. $100 to $200 [29] for Protein Total Pleural Fluid (pleural protein), $40 to $70 [30] for c-reactive protein test, and $6 to $167 [31] for complete blood count (platelet count) depending upon the location. These factors can also advocate early prognosis of MM; Moreover, studies have shown that higher (>1 mg/dL) c-reactive protein influences mesothelioma [32] [33], another study at the University of Maryland determined the clinical significance of preoperative thrombocytosis (high count of platelets), in patients with MPM [34].

## 5. Conclusion

Our study identifies that the diagnosis method (Biopsy and imaging test results), c-reactive protein, platelet count, duration of symptoms, gender, and pleural protein plays a significant role in diagnosing PM. However, effective diagnosis methods such as pleuroscopy (lungs) or laparoscopy (abdomen), thoracotomy (lungs) or laparotomy (abdomen), and imaging tests (CT scan and MRI) are expensive. This study proposes two approaches to predict PM, each having its advantages and limitations. The first approach (phase 1) uses all predictors from mesothelioma data and produces 100% classification accuracy. The second approach (phase 2) ensures cost reduction. Our study recommends AdaBoost algorithms for PM prognosis and suggests using the phase-2 approach to shortlist high-risk patients, followed by phase 1 to confirm PM.

## 6. List of abbreviations

- MPM — Malignant Pleural Mesothelioma
- MM — Malignant Mesothelioma
- PM — Pleural Mesothelioma
- ROC — Receiver Operating Characteristics
- PRC — Precision-recall curve
- DT — Decision tree
- VFDT — Very fast decision tree
- MLP — Multi-layer perceptron
- SGD — Stochastic gradient descent
- KLR – Kernel logistic regression
- AdaBoost — Adaptive boosting
- RMSE — Root mean squared error
- ANN — Artificial neural network
- SVM — Support vector machine
- S-Pegasos — Primal Estimated sub-Gradient Solver for SVM
- CC — Clojure classification
- VP — Voted perceptron
- TP — True positive
- TN — True negative
- FP — False positive
- FN — False negative
- TPR — True positive rate
- FPR — False positive rate
- WBC — White blood cell
- HGB – Hemoglobin
- PLT — Platelet count
- LDH — Blood lactic dehydrogenize
- ALP — Alkaline phosphate
- CRP — C reactive protein

- AUC — Area under the curve

## 7. Declarations

- Competing interests — The authors declare that they have no competing interests
- Funding — Any internal or external source did not fund this study

## References

1. P. Courtiol, C. Maussion, M. Moarii et al., Deep learning-based classification of Mesothelioma improves prediction of patient outcome., *Nature Medicine* 25(1) (2019), 1519–1525. doi:10.1038/s41591-019-0583-3.

2. N.J. Vogelzang, J.J. Rusthoven, J. Symanowski, C. Denham, E. Kaukel, P. Ruffie, U. Gatzemeier, M. Boyer, S. Emri, C. Manegold, C. Niyikiza, and P. Paoletti, Phase III Study of Pemetrexed in Combination With Cisplatin Versus Cisplatin Alone in Patients With Malignant Pleural Mesothelioma, *Journal of Clinical Oncology* 21(14) (2003), 2636–2644. doi:10.1200/jco.2003.11.136.

3. G. Zalcman, J. Mazieres, J. Margery, L. Greillier, C. Audigier-Valette, D. Moro-Sibilot, O. Molinier, R. Corre, I. Monnet, V. Gounant, F. Rivière, H. Janicot, R. Gervais, C. Locher, B. Milleron, Q. Tran, M.-P. Lebitasy, F. Morin, C. Creveuil, J.-J. Parienti and A. Scherpereel, Bevacizumab for newly diagnosed pleural Mesothelioma in the Mesothelioma Avastin Cisplatin Pemetrexed Study (MAPS): a randomised, controlled, open-label, phase 3 trial, *The Lancet* 387(10026) (2016), 1405–1414. doi:10.1016/s0140-6736(15)01238-6.

4. H. Pass, D. Giroux, and C. Kennedy, The IASLC mesothelioma staging project: improving staging of a rare disease through 29. international participation, *J Thorac Oncol* 11 (2016), 2082–2088.

5. R. Gill, D. Naidich and A. Mitchell, North American multicenter volumetric CT study for clinical staging of malignant pleural 30. mesothelioma: feasibility and logistics of setting up a quantitative imaging study, *J Thorac Oncol* 11 (2016), 1335–1344.

6. T. Frauenfelder, M. Tutic, W. Weder, R.P. Gotti, R.A. Stahel, B. Seifert, and I. Opitz, Volumetry: an alternative to assess therapy response for malignant pleural Mesothelioma? *European Respiratory Journal* 38(1) (2011), 162–168. doi:10.1183/09031936.00146110.

7. R. Bueno, E. Stawiski, and L. Goldstein, Comprehensive genomic analysis of malignant pleural Mesothelioma identifies recurrent mutations, gene fusions, and splicing alterations, *Nat Genet* 48 (2016), 407–416.

8. J.C. McDonald and A.D. McDonald, The epidemiology of Mesothelioma in historical context, *European Respiratory Journal* 9(9) (1996), 1932–1942. doi:10.1183/09031936.96.09091932.

9. C.W. Noonan, Environmental asbestos exposure and risk of Mesothelioma, *Annals of translational medicine* 5(11) (2017), 234. doi:10.21037/atm.2017.03.74.

10. S.J. Henley, T.C. Larson, M. Wu, V.C.S. Antao, M. Lewis et al., Mesothelioma incidence in 50 states and the District of Columbia, United States, 2003–2008, *International Journal*

*of occupational and environmental health* 19(1) (2013), 1–10. doi:10.1179/2049396712Y.0000000016.

11. J. Leigh, C.F. Corvalán, A. Grimwood, G. Berry, D.A. Ferguson, and R. Thompson, The incidence of malignant Mesothelioma in Australia 1982–1988, *American Journal of Industrial Medicine* 20(5) (1991), 643–655. doi:10.1002/ajim.4700200507.

12. M. News, What Mesothelioma Does to the Body, 2018. http://www.mesotheliomanews.com/medical/mesotheliomadiagnosis/pleural-mesothelioma/.

13. K. Selby, Mesothelioma Diagnosis, 2018. https://www.asbestos.com/mesothelioma/diagnosis/.

14. L. Molinari, Mesothelioma Treatment Costs, 2018. https://www.mesothelioma.com/treatment/mesothelioma-treatmentcosts/.

15. R.B. Friedin, Am I going to die because I cannot afford the test?, 2012. https://www.kevinmd.com/blog/2012/05/dieafford-test.html.

16. T.P. Pope, When Patients Can't Afford Their Care, 2010. https://well.blogs.nytimes.com/2010/02/04/when-patients-cantafford-their-care/.

17. J. Silvester, Most of the World Doesn't Have Access to X-Rays., 2016. https://www.theatlantic.com/health/archive/2016/ 09/radiology-gap/501803/.

18. H.O. Ilhan and E. Celik, The mesothelioma disease diagnosis with artificial intelligence methods, in *IEEE 10th International Conference on Application of Information and Communication Technologies*, 2016.

19. K. Tin, Random decision forests, *Proceedings of 3rd International Conference on Document Analysis and Recognition* (1995).

20. E. Lotfi and A. Keshavarz, Gene expression microarray classification using PCA–BEL, *Computers in Biology and Medicine* 54 (2014), 180–187. doi:10.1016/j.compbiomed.2014.09.008. https://dx.doi.org/10.1016/j.compbiomed.2014.09.008.

21. M. Kuhn and K. Johnson, Applied Predictive Modeling, Springer, New York, 2018. doi:10.1007/978-1-4614-6849-3. https://link.springer.com/content/pdf/10.1007/978-1-4614-6849-3.pdf.

22. G. James, D. Witten, T. Hastie, and R. Tibshirani, An Introduction to Statistical Learning: with Applications in R, et al., ed., Springer, 2017. doi:10.1007/978-1-4614-7138-7. https://link.springer.com/content/pdf/10.1007/978-1-4614-7138-7.pdf.

23. M.B. Kursa and W.R. Rudnicki, Feature Selection with theBorutaPackage, *Journal of Statistical Software* 36(11) (2010), 2–13. doi:10.18637/jss.v036.i11.

24. R. Nilsson, J.M. Peña, J. Björkegren, and J. Tegner, Consistent Feature Selection for Pattern Recognition in Polynomial' Time, *The Journal of Machine Learning Research* 8 (2007), 612–612.

25. A. Liaw and M. Wiener, Classification and Regression by random, *Forest. R News* 2(3) (2002), 18–22.

26. S.-J. Lee, T. Chen, L. Yu, and C.-H. Lai, Image Classification Based on the Boost Convolutional Neural Network, *IEEE Access* 6 (2018), 12755–12768.

27. R. Chojniak, Incidentalomas: managing risks, *Radiologia brasileira* 48(4) (2015), 9–10. doi:10.1590/01003984.2015.48.4e3.

28. V. Ascoli, Pathologic diagnosis of malignant Mesothelioma: Chronological prospect and advent of recommendations and guidelines, *Ann Ist Super Sanità* 51(1) (2015), 52–59.

29. Practo, Protein Total Pleural Fluid, 2017. https://www.practo.com/tests/protein-total-pleural-fluid/p.

30. M. Haiken, 3 New Medical Tests that Can Save Your Life - But You Have to Ask, 2011. https://www.forbes.com/sites/ melaniehaiken/2011/06/17/3-lifesaving-new-medical-tests-you-have-to-ask-for/#2df47f75398a.

31. J. Pinder, How much does a blood test cost? It could be $6, or $167. (Clear Health Cost Beta), 2012. https://clearhealthcosts.com/blog/2012/12/how-much-does-a-blood-test-cost-it-could-be-16-or-117/.

32. S. Takamori, G. Toyokawa, M. Shimokawa, F. Kinoshita, Y. Kozuma, T. Matsubara, N. Haratake, T. Akamine, F. Hirai, T. Seto, T. Tagawa, M. Takenoyama, Y. Ichinose, and Y. Maehara, The C-Reactive Protein/Albumin Ratio is a Novel Significant Prognostic Factor in Patients with Malignant Pleural Mesothelioma: A Retrospective Multi-institutional Study, *Annals of Surgical Oncology* 25(6) (2018), 1555–1563. doi:10.1245/s10434-018-6385-x.

33. B. Ghanim, M.A. Hoda, M.-P. Winter, T. Klikovits, A. Alimohammadi, B. Hegedus, B. Dome, M. Grusch, M. Arns, P. Schenk, W. Pohl, C. Zielinski, M. Filipits, W. Klepetko and W. Berger, Pretreatment Serum C-Reactive Protein Levels Predict Benefit From Multimodality Treatment Including Radical Surgery in Malignant Pleural Mesothelioma, *Annals of Surgery* 256(2) (2012),357–362. doi:10.1097/sla.0b013e3182602af4.

34. Y.C. Li, T. Khashab, J. Terhune, R.L. Eckert, N. Hanna, A. Burke, and H.R. Alexander, Preoperative Thrombocytosis Predicts Shortened Survival in Patients with Malignant Peritoneal Mesothelioma Undergoing Operative Cytoreduction and Hyperthermic Intraperitoneal Chemotherapy, *Annals of Surgical Oncology* 24(8) (2017), 2259–2265. doi:10.1245/s10434017-5834-2.